\documentclass[letterpaper, 10 pt, conference]{IEEEtran}

\IEEEoverridecommandlockouts

\usepackage{epsfig}
\usepackage{amsmath}
\usepackage{amssymb}
\usepackage{gensymb}
\usepackage[T1]{fontenc}
\usepackage{algpseudocode}
\usepackage{mathtools}
\usepackage{multirow}
\usepackage{makecell}
\usepackage{bm}
\usepackage{algorithm}

\usepackage{epstopdf}
\usepackage{pifont}
\usepackage{multirow}
\usepackage{url}
\usepackage{tabularx}
\usepackage[linkcolor=black,citecolor=black,urlcolor=black,colorlinks=true]{hyperref}
\usepackage[skip=3pt, font={small}]{caption}
\usepackage{booktabs}
\usepackage{array}
\usepackage[dvipsnames]{xcolor}
\newcommand{\cmark}{\textcolor{BrickRed}{\textbf{\large \ding{51}}}}
\newcommand{\authorrefmark}[1]{%
\ifcase#1\relax
\or\textsuperscript{*}
\or\textsuperscript{\dagger}
\fi
}

\title{\LARGE \bf
CoRe-MoE: Contrastive Reweighted Mixture of Experts for Multi-Terrain Humanoid Locomotion with Gait Adaptation}

\author{
    Kailun Huang$^{1,*}$\quad
    Zikang Xie$^{1,*}$\quad
    Yanzhe Xie$^{1,*}$\quad
    Panpan Liao$^{4,*}$\quad
    Fanghai Zhang$^{1}$\quad
    Yanheng Mai$^{1}$\\

    Wenhao Xu$^{3}$\quad
    Yunheng Wang$^{1}$\quad
    Renjing Xu$^{1,\dagger}$\quad
    Haohui Huang$^{4,\dagger}$\quad
    Chenguang Yang$^{2,\dagger}$~\IEEEmembership{Fellow,~IEEE}\\

    $^1$Hong Kong University of Science and Technology (Guangzhou)\quad
    $^2$Hong Kong Polytechnic University\\
    $^3$South China Agricultural University\quad
    $^4$Guangdong University of Technology\\

    $^{*}$ Equal contribution \quad
    $^{\dagger}$ Corresponding authors
}

\makeatletter
\IEEEaftertitletext{%
  \vspace{-0.9cm}
  \begin{center}
    \includegraphics[width=\linewidth]{figure/cover.jpg}
    \captionof{figure}{\label{figure:cover}We propose \textbf{CoRe-MoE} enables smooth, natural, and continuous transitions between walking and running using a single unified policy, while maintaining stable adaptability to complex terrain. In contrast, \textbf{Hiking in the Wild} relies on separately trained walking and running policies that must be manually switched during execution, leading to abrupt transitions, reduced continuity, and noticeably unnatural motion, especially under terrain variations. \textbf{Project Page: \href{https://core-moe.github.io/}{https://core-moe.github.io/}.}}
    \vspace{-0.15cm}
  \end{center}
}
\makeatother

\begin{document}

\maketitle
\thispagestyle{empty}
\pagestyle{empty}

\begin{abstract}
Humans primarily rely on walking and running to traverse complex terrains. Similarly, humanoid robots should be able to smoothly transition between walking and running while maintaining natural and stable locomotion. However, unifying gait transition and multi-terrain adaptation within a single policy remains challenging due to gradient interference between tasks and the distribution shift caused by terrain variations. Although Mixture-of-Experts (MoE) architectures can mitigate multi-skill interference, direct joint training often fails to achieve clear expert specialization. To address these challenges, we propose CoRe-MoE, a two-stage reinforcement learning framework that decouples gait generation from terrain adaptation. In the first stage, a stable locomotion policy is learned to produce natural walking and running behaviors with smooth transitions. In the second stage, a terrain-aware MoE branch is introduced, and the gating network is trained with a contrastive objective to learn structured terrain representations and promote expert specialization. The final action is obtained through weighted fusion of the base gait policy and the terrain-aware branch, enabling the policy to preserve stable locomotion while adapting to complex terrains. Extensive simulation results demonstrate that the proposed method outperforms baseline approaches in terms of success rate, locomotion stability, and multi-terrain adaptability. Furthermore, zero-shot deployment on a Unitree G1 humanoid robot validates the effectiveness of our framework, achieving robust walking and running across stairs, slopes, steps, obstacles, and unstructured outdoor terrains while maintaining accurate foothold control and dynamic stability.
\end{abstract}


\section{Introduction}
\IEEEPARstart{H}{umanoid} robots are increasingly required to perform tasks in real-world environments, which often contain abrupt terrain variations such as stairs, slopes, and steps. Compared with wheeled robots \cite{PIE, dreamwaq++}, humanoids can step over obstacles and traverse discontinuous terrains, making them more suitable for deployment in human-centered environments \cite{zhu2026hiking, 2024realrlloco, loco1, beamdojo}. To achieve robust and human-like locomotion in such scenarios, robots must not only maintain basic stability but also proactively adjust their actions according to upcoming terrain changes \cite{PIM, long2024hybrid}.

Humans rely primarily on two fundamental gaits: walking and running to adapt to complex terrains \cite{gait1, gait2}, rarely employing additional complex or ineffective gaits to cross stairs, slopes, or gaps. This observation provides a design insight for humanoid robots: control strategies should focus on smooth switching between walking and running, rather than learning numerous artificial gait types\cite{MORE}, which contribute little to practical adaptability.

Despite the significant progress in reinforcement learning–based humanoid locomotion, key challenges remain. Pure proprioception–based “blind locomotion” methods can resist moderate disturbances through contact feedback \cite{humanoid_gym, blind_bipedal, gu2024, radosavovic2024}, but their reactive nature makes it difficult for the robot to anticipate hazards such as high steps or deep gaps. On the one hand, this can lead to instability or falls on critical terrains; on the other hand, jointly achieving gait switching and multi-terrain adaptation within a single policy often induces gradient conflicts, while the visual and dynamic variations introduced by complex terrains further increase learning difficulty.

In addition, Mixture-of-Experts (MoE) architectures \cite{MOE} have shown promising potential in multi-skill learning and have been gradually applied to humanoid locomotion and multi-terrain adaptation \cite{moe_loco, MORE}. However, during direct training, the experts often fail to develop clear specialization: their activation patterns tend to be dispersed and highly overlapping across different terrains or gait conditions, preventing the policy from fully exploiting the advantages of the MoE structure and limiting its multi-terrain adaptability and gait control precision.

To address these issues, we propose the Contrastive Reweighted Mixture of Experts (CoRe-MoE), a two-stage training framework that integrates adaptive gait policies, visual perception, and contrastive MoE learning. Depth images \cite{zhu2026hiking, zhuang2026deepwholebodyparkour} are incorporated to enable proactive, terrain-aware motion adjustment. In the first stage, the policy learns stable walking and running on flat terrains and acquires natural and smooth gait transition capabilities. In the second stage, we introduce a terrain-aware MoE branch guided by SwAV-style contrastive learning \cite{swav}. This encourages the network to learn structured terrain representations and promotes expert specialization. The final action is produced by a weighted fusion of the base gait policy and the terrain-aware branch, enabling robust multi-terrain adaptability while preserving stable fundamental gait behaviors.

As illustrated in Figure \ref{figure:cover}, CoRe-MoE achieves continuous, natural transitions between walking and running within a single policy while maintaining stable performance on complex terrain. In contrast, existing methods—such as Hiking in the Wild—train walking and running separately and require manual switching during execution, leading to abrupt transitions, poor continuity, and degraded stability when terrain conditions change.

Simulation results demonstrate that CoRe-MoE outperforms baseline methods in terms of success rate, gait stability, and multi-terrain adaptability. Further zero-shot deployment experiments on the Unitree G1 robot show that the policy achieves robust locomotion across stairs, slopes, and outdoor natural terrains, while maintaining accurate foot placement and dynamic stability under external disturbances, highlighting its potential for real-world applications.

The main contributions of this work are summarized as follows:

\begin{itemize}
    \item We propose CoRe-MoE, a two-stage end-to-end training framework that enables smooth switching between walking and running while achieving multi-terrain adaptability and robustness.
    \item We design a terrain-aware MoE module based on contrastive learning that generates terrain-conditioned action adjustments and promotes clear expert specialization in different terrains.
    \item We perform a zero-shot deployment on the Unitree G1 robot, and the results demonstrate robust locomotion and dynamic stability across stairs, slopes, and natural outdoor terrains.
\end{itemize}

\begin{table}[h]
    \centering
    \renewcommand{\arraystretch}{1.0}
    \setlength{\tabcolsep}{4pt}
    \fontsize{8pt}{10pt}\selectfont
    \caption{Comparison of locomotion capabilities across different methods. Unlike prior approaches that support limited gaits or require policy switching, CoRe-MoE is the first unified policy achieving walking, running, smooth transitions, and robust multi-terrain performance.}
    \label{table:Comparison}
    \begin{tabular}{
        >{\centering\arraybackslash}p{2.8cm}  
        >{\centering\arraybackslash}p{0.5cm}  
        >{\centering\arraybackslash}p{0.5cm}  
        >{\centering\arraybackslash}p{1.8cm}  
        >{\centering\arraybackslash}p{1.8cm}
    }
        \toprule
        \textbf{Method} & \textbf{Walk} & \textbf{Run} & 
        \textbf{\mbox{Auto Transition}} & 
        \textbf{\mbox{Terrain Adapt}} \\
        \midrule
        Beamdojo \cite{beamdojo}          & \ding{51} & \ding{55} & \ding{55} & \ding{51} \\
        Now You See That \cite{sun2026thatlearningendtoendhumanoid} & \ding{51} & \ding{55} & \ding{55} & \ding{51} \\
        MoRE \cite{MORE}              & \ding{51} & \ding{51} & \ding{55} & \ding{51} \\
        Hiking in the Wild \cite{zhu2026hiking} & \ding{51} & \ding{51} & \ding{55} & \ding{51} \\
        \textbf{CoRe-MoE} & \cmark & \cmark & \cmark & \cmark \\  \bottomrule
    \end{tabular}
\vspace{-0.3cm}
\end{table}

\begin{figure*}
    \centering
    \includegraphics[width=1.0\linewidth]{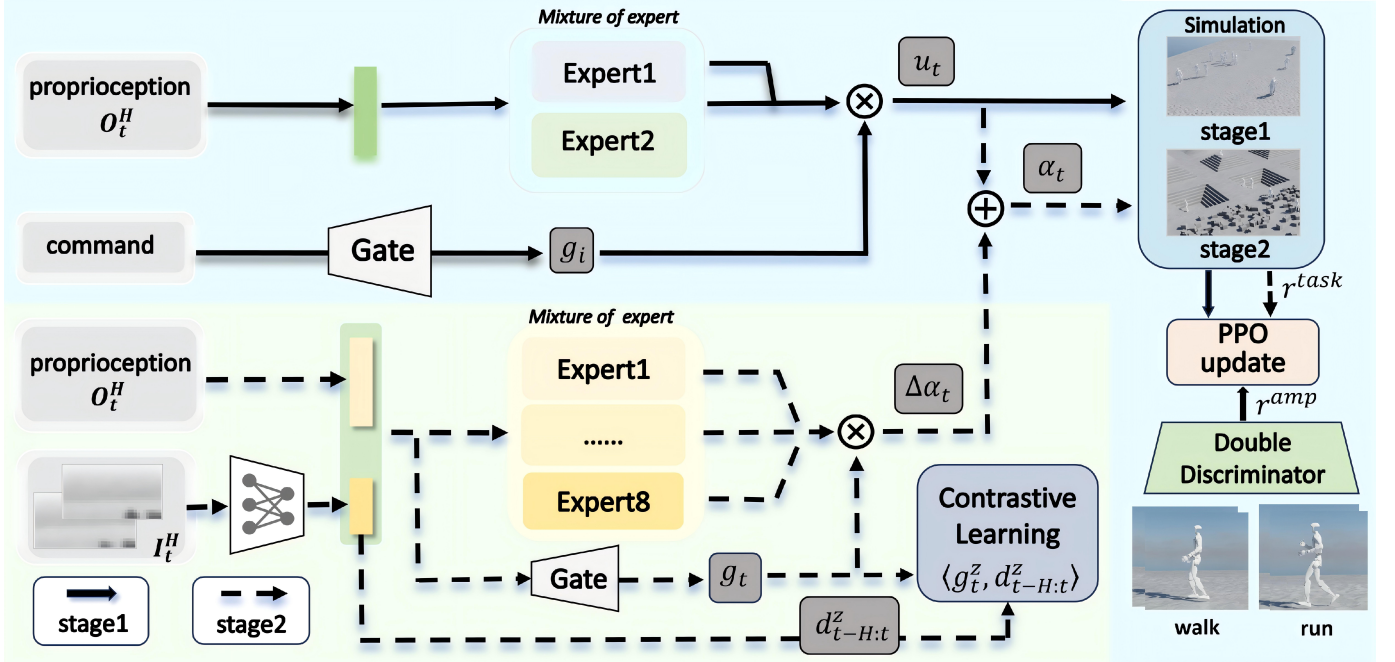}
\caption{Overview of our framework. In Stage 1, we learn a flat-terrain adaptive gait policy using MoE and AMP, enabling stable walking–running transitions. In Stage 2, we incorporate depth perception and employ MoE with contrastive learning to acquire terrain-aware control. During inference, the flat-terrain policy provides a gait prior and the multi-terrain policy applies terrain adjustments; their weighted fusion produces the final action, achieving natural gait and multi-terrain adaptability.}
\vspace{-0.3cm}
\label{figure:pipeline}
\end{figure*}

\section{Related Work}
\subsection{Learning-based Humanoid Robot Control}
Reinforcement learning (RL) has recently achieved substantial progress in humanoid locomotion. RL-based policies can learn stable walking and running behaviors directly from proprioceptive feedback and maintain robustness on moderately challenging terrains \cite{radosavovic2024real, siekmann2021blind, humanoid_gym, yuan2025pvp}. Compared to traditional model-based controllers \cite{HumanoidMPC, scianca2020mpc, mpc}, these learning-based approaches are capable of handling unknown dynamics and unstructured environments, leading to significantly improved adaptability.

Although prior works have explored multi-gait or multi-terrain locomotion \cite{MORE, moe_loco, gait_adaptive}, their policies often rely on manually triggered gait switches and cannot autonomously adjust gait according to speed or terrain commands. Their generalization across fundamentally different terrain types such as stairs, gaps, and obstacles also remains limited, as shown in Table~\ref{table:Comparison}.

Human locomotion studies show that people primarily rely on only two fundamental gaits—walking and running—to traverse diverse terrains, without resorting to additional artificial or inefficient gait patterns. This observation suggests that practical humanoid locomotion should emphasize autonomous and smooth transitions between these two core gaits while simultaneously maintaining strong multi-terrain adaptability.

\subsection{Perception for Locomotion}
In recent years, researchers have incorporated exteroceptive perception into humanoid locomotion policies \cite{sun2026thatlearningendtoendhumanoid,cheng2024extreme,PIE,beamdojo,rudin2025parkour}to enhance proactive adaptation to complex terrains. Blind locomotion methods, which rely solely on proprioceptive signals, can maintain stability on limited terrains but cannot anticipate upcoming terrain features. Traditional LiDAR-based approaches can provide height information \cite{he2025attention, PIM, ben2025gallant}, but their limited field-of-view, especially under the robot’s feet, combined with state estimation drift, low update frequency, and motion distortion, restricts their effectiveness in high-speed or dynamic scenarios~\cite{perceptive_internal_model,beamdojo}. In contrast, depth cameras provide high-frequency perception and can be integrated with policy networks for end-to-end control\cite{zhu2026hiking, zhuang2024humanoid, zhuang2026deepwholebodyparkour}, enabling the robot to proactively adjust its actions to handle complex terrains, thereby improving safety and adaptability.

\subsection{Multi-terrain Adaptation and Mixture-of-Experts}
Mixture-of-Experts architectures \cite{MOE, 6796382} offer an effective approach for multi-skill learning and multi-terrain adaptation. By dividing the policy into specialized experts, the robot can learn distinct skills such as walking, running, and gap-crossing\cite{MORE,moe_loco}. However, during direct training, expert activations are often dispersed and lack specialization, limiting the policy’s ability to adapt across different terrains. Contrastive learning methods, such as SwAV \cite{swav} and SimCLR \cite{simclr}, can guide experts to specialize according to terrain conditions, enhancing both policy robustness and multi-terrain adaptability. Our proposed CoRe-MoE framework combines weighted action correction with contrastive MoE learning, preserving gait stability while enabling proactive adaptation and flexible control over complex terrains.

\section{Method}
To address the coupled challenges of terrain disturbances and gait transitions, we aim for humanoid robots to maintain stable dynamics while achieving natural and continuous gait switching in complex environments. We adopt a two-stage training pipeline: the first stage focuses on learning stable and controllable fundamental gaits such as walking and running; the second stage extends the objective to enable multi-terrain adaptation while preserving a consistent anthropomorphic motion style, as illustrated in Figure \ref{figure:pipeline}. This staged design improves training stability and allows the robot to perform smooth and coherent gait transitions across diverse terrains.

\subsection{Problem Description}
We formulate humanoid multi-terrain locomotion as a Partially Observable Markov Decision Process (POMDP) with a tuple  $\langle S, A, P, R, \gamma \rangle$, where $S$ is the state space, $A$ is the action space,  $P(|s,a)$ is the transition function. The agent receives a reward from $R(s, a)$. The discount factor $\gamma \in [0, 1]$ balances immediate and future rewards. We use Proximal Policy Optimization (PPO) \cite{ppo} to learn the optimal policy $\pi^*$ that maximizes the expected discounted return:
\begin{equation}
\pi^* = \arg \max \mathbb{E}\left[\sum\limits_{t = 0}^\infty \gamma^t R(s_t, a_t)\right].
\end{equation}
We adopt an asymmetric actor–critic architecture. The policy network relies solely on deployable, non-privileged proprioceptive observations, defined at time t as:
\begin{equation}
O_t = [\omega_t, g_t, c_v^t, \theta_t, \dot{\theta}_t, a_{t-1}], 
\end{equation}
including robot angular velocity $\omega_t$, gravity direction $g_t$, velocity command $c_v^t$, joint angle $\theta_t$, velocity $\dot{\theta}_t$, last action $a_{t-1}$. In addition, the policy also receives the perception observation depth image $I_t$.
The critic receives the same observation as the actor but is additionally provided with privileged base linear velocity $v_t$.

\subsection{Adaptive Gait Locomotion Policy}
In the first training stage, we aim to learn a stable and controllable locomotion policy that enables the humanoid robot to adaptively switch between walking and running on flat terrain. We employ two separate Adversarial Motion Priors (AMP) \cite{amp} networks, one for walking and one for running. Additionally, the walking trajectories are augmented with gait samples such as stair climbing and obstacle crossing to further enhance gait diversity and adaptability.

The policy adopts MoE architecture with historical frame inputs $O_t^H$ to capture temporal dependencies. It is important to note that the depth image $I_t$ is not used in this stage; locomotion control on flat terrain relies solely on proprioceptive observations to achieve gait switching. The MoE gating network takes the commanded velocity $c_v^t$ as input and dynamically activates different experts, enabling the walking AMP under low-speed commands and the running AMP under high-speed commands. The final action is computed as a weighted sum of expert outputs through softmax gating:
\begin{equation}
u_t = \sum_{i=1}^{N} \text{softmax}({g}_i) \cdot u_i,
\end{equation}
where $z_i$ means the $i$-th expert outputs and ${g}_i$ is the expert activation.

The training reward consists of three terms, combined as
\begin{equation}
\ r_t = r_{task} + r_{amp} + r_{reg} , 
\end{equation}
where $r_{amp}$ is provided by the activated AMP network to enforce realistic walking or running motions, $r_{task}$ ensures locomotion stability and that the commanded velocity is achieved, $r_{reg}$ is a regularization term that penalizes large or abrupt joint motions to promote smooth and physically plausible actions. By optimizing this combined reward, the policy learns to smoothly transition between walking and running, achieving natural and continuous gait adaptation while maintaining dynamic stability on flat terrain. After completing Stage 1 training, the entire policy network is frozen.

\subsection{Adaptive Multi-Terrain Policy}
In the second training stage, we introduce a multi-terrain policy on top of the adaptive gait policy learned in Stage 1 to enhance locomotion performance on complex terrains. This policy is built using a MoE architecture, where each expert receives not only historical proprioceptive observations $O_t^H$ but also the corresponding historical depth image sequence $I_t^H$, allowing the model to capture both temporal dynamics and external terrain geometry. During this stage, the Stage 1 policy network is frozen, and the actions generated by the Stage 2 policy are added to those from Stage 1, enabling terrain-aware fine-grained adjustments while preserving the stable gait foundation learned previously. The reward function in this stage retains only the task reward $r_{task}$ and regularization reward $r_{reg}$, ensuring the policy focuses on terrain adaptation while maintaining smooth and physically consistent actions.

To fully activate experts and encourage terrain-specific specialization, we adopt a SwAV-based contrastive learning objective that aligns the gating output with the encoded representation of the history of the depth image. Compared with single-frame information, the historical depth sequence provides more stable and structured cues about terrain geometry, enabling the gating network to form clearer terrain clusters. We align the gating embedding $g_t^z$ with the depth history representation $d_{t-H:t}^z$
\begin{equation}
\langle g_t^z, d_{t-H:t}^z \rangle
\end{equation}
During each iteration, we collect batch trajectories from all environments. If a historical observation and its corresponding historical depth sequence belong to the same trajectory, they form a positive pair; otherwise, they are negative. The gate network output $g_t$ serves as the source vector $I_{\text{source}}^{t}$, while the encoded representation of the historical depth sequence $d_{t-H:t}^{z}$ serves as the target vector $I_{\text{target}}^{z}$. Both vectors are normalized before computing the contrastive objective.

To predict the cluster assignment probabilities, we first normalize the prototype matrix:
\begin{equation}
\mathbf{E}=\{ \bar{\mathbf{e}}_1, ..., \bar{\mathbf{e}}_K \},
\end{equation}
and compute the source and target probabilities:
\begin{equation}
\begin{aligned}
    \mathbf{p}^{t}_\text{source} &= 
    \frac{\exp(\frac{1}{\tau} {I^t_\text{source}}^\top\mathbf{e}_k)}
         {\sum_{k'}\exp(\frac{1}{\tau} {I^t_\text{source}}^\top\mathbf{e}_{k'})} \\
    \mathbf{p}^{t}_\text{target} &= 
    \frac{\exp(\frac{1}{\tau} {I^t_\text{target}}^\top\mathbf{e}_k)}
         {\sum_{k'}\exp(\frac{1}{\tau} {I^t_\text{target}}^\top\mathbf{e}_{k'})}
\end{aligned}
\end{equation}
where $\tau$ is a temperature parameter. To avoid trivial solutions, the Sinkhorn-Knopp algorithm is applied to both source and target probabilities to obtain target distributions $q_\text{target}^t$ $q_\text{source}^t$.
The final SwAV contrastive loss is:
\begin{equation}
\mathcal{J} = -\frac{1}{2H} \sum^{H}_{t=1} (\mathbf{q}^t_\text{target} \log \mathbf{p}^t_\text{source} + \mathbf{q}^t_\text{source} \log \mathbf{p}^t_\text{target}).
\end{equation}

By aligning the embedding of the gating with the historical depth representation, the SwAV objective enables the gating network to better discriminate variations in terrain. As a result, different terrain types induce distinct cluster assignments, which in turn encourage the MoE experts to develop specialized behaviors. With more accurate expert activation, the policy is able to produce terrain-aware corrections rather than generic modifications. The MoE-weighted action:
\begin{equation}
\Delta a_t = \sum_{i=1}^{N} \text{softmax}({g}_t) \cdot a_i,
\end{equation}
which reflects the specific adjustments required by the upcoming terrain.

\begin{figure}
    \centering
    \includegraphics[width=1\linewidth]{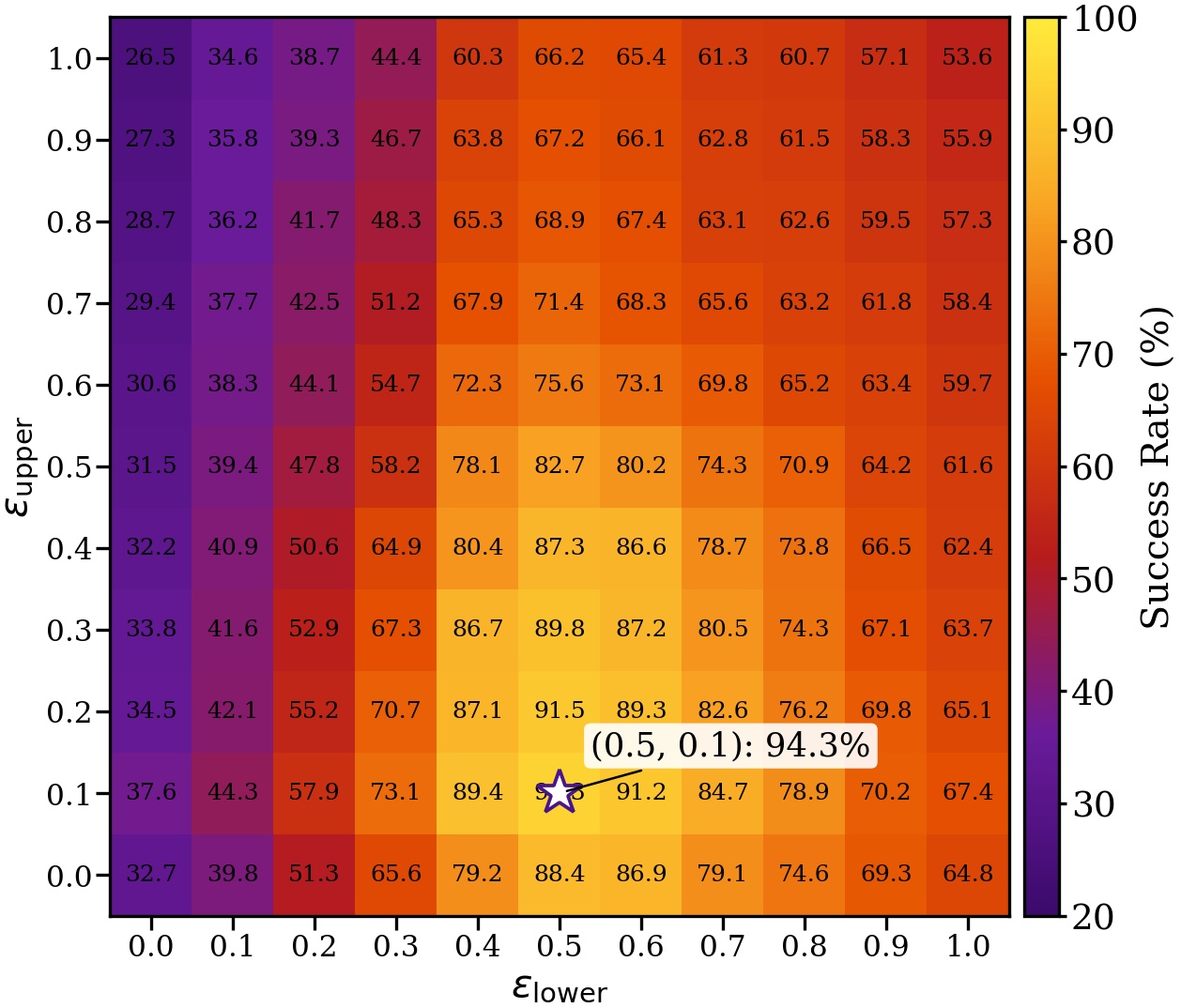}
    \caption{Sensitivity analysis of the action fusion coefficients $\epsilon_{lower}$ and $\epsilon_{upper}$. Colors indicate the success rate under multi-terrain evaluation.}
    \vspace{-0.5cm}
    \label{figure:sensitivity}
\end{figure}

For action fusion, we apply different blending coefficients to different body parts to control the influence of terrain-aware corrections: $\epsilon_{upper} = 0.1$ for the upper body and $\epsilon_{lower} = 0.5$ for the lower body. The final action is defined as:
\begin{equation}
 a_t = u_t + \epsilon \cdot \Delta a_t,
\end{equation}
where $\epsilon$ is applied per body region accordingly. This design preserves the stable gait structure learned in Stage 1 while adding fine-grained, terrain-dependent refinements. 

To investigate the influence of the fusion coefficients, we further performed a sensitivity analysis on $\epsilon_{upper}$ and $\epsilon_{lower}$, as shown in Fig.~\ref{figure:sensitivity}. The results indicate that the policy is more sensitive to $\epsilon_{lower}$, highlighting the dominant role of terrain-aware corrections of the lower-body in the adaptation of the foothold and the traversal of the terrain. When $\epsilon_{lower}$ is too small, the policy cannot respond effectively to terrain variations, whereas excessively large values may introduce overly aggressive corrections that compromise gait stability. In contrast, $\epsilon_{upper}$ has a relatively smaller impact on overall performance. The results show that the adopted configuration lies within a high-performance region, indicating that the proposed action fusion mechanism achieves a favorable balance between locomotion stability and terrain adaptability.

Through this mechanism, the multi-terrain branch serves as a lightweight yet effective adaptation layer, driven by contrastively shaped gating behavior, enabling robust locomotion across diverse and uneven terrains.

\section{Experiment Configurations}
\subsection{Training Parameters}
We train all policies in NVIDIA Isaac Sim and Isaac Lab \cite{isaaclab} 4096 parallel environments to ensure stable gradient estimates and rapid throughput. Motion priors are constructed from selected AMASS \cite{amass} and LAFAN \cite{lafan} sequences, retargeted to the Unitree G1 model and used to supervise the AMP discriminator. The first-stage adaptive gait policy is trained for 25k iterations on an RTX 4090 GPU with a 2-expert MoE, while the second-stage residual multi-terrain policy is trained for 20k iterations with the MoE capacity expanded to 8 experts. For the contrastive learning module, we set the number of prototypes at 16 and the temperature at 0.1. Depth image preprocessing, including downsampling and spatial cropping, follows the procedure referenced in \cite{zhu2026hiking}, ensuring consistent perceptual inputs across simulation and real-world deployment.

\subsection{Reward Functions}
Our reward formulation is based on standard locomotion objectives while incorporating two essential safety considerations. The first focuses on preventing missteps in unsupported terrain regions \cite{beamdojo}, ensuring that the policy avoids foot placements that could lead to immediate loss of balance. The second integrates a scalable terrain-edge detection module with foot-volume contact evaluation \cite{zhu2026hiking}, encouraging the robot to land within a safe foothold margin and reducing the likelihood of edge-related slips. A complete description of all reward terms is provided in Table \ref{table:reward}.

\begin{table}[!htbp]
\centering
\fontsize{8pt}{6pt}\selectfont 
\caption{Reward Functions for Humanoid Locomotion}
\label{table:reward}

\begin{tabularx}{\columnwidth}{
    @{}
    >{\raggedright\arraybackslash}p{2.3cm}
    >{\raggedright\arraybackslash}p{4.2cm} 
    >{\centering\arraybackslash}p{0.8cm}
    @{}
}
\toprule
\textbf{Term} & \textbf{Equation} & \textbf{Weight} \\
\midrule

Velocity tracking & $\exp(-\|\mathbf{v}_{xy}-\mathbf{v}_{xy}^c\|_2^2/\sigma^2)$ & $2.0$ \\
Yaw tracking & $\exp(-\|\omega_z-\omega_z^c\|_2^2/\sigma^2)$ & $2.0$ \\
Heading error & $\|\psi-\psi^c\|_2^2$ & $-1.0$ \\
Is alive & $1$ & $3.0$ \\
Stand still & $\mathbb{I}(\|\mathbf{v}^c\|<\epsilon)\,\|\mathbf{v}\|_2^2$ & $-0.3$ \\
Don't wait & $\mathbb{I}(\|\mathbf{v}^c\|>\epsilon)\,\mathbb{I}(\|\mathbf{v}\|<\delta)$ & $-0.5$ \\
\midrule
Orientation & $\|\mathbf{g}_{xy}\|_2^2$ & $-3.0$ \\
Pelvis orientation & $\|\mathbf{q}_{\text{err,pelvis}}\|_2^2$ & $-3.0$ \\
Angular velocity & $\|\boldsymbol{\omega}_{xy}\|_2^2$ & $-0.05$ \\
Joint deviation hip & $\sum_{i\in\text{hip}}(\theta_i-\theta_i^{\text{def}})^2$ & $-0.5$ \\
Freeze upper body & $\sum_{i\in\text{upper}}|\theta_i-\theta_i^{\text{def}}|$ & $-0.004$ \\
Feet air time & $\sum_i t_{\text{air},i}$ & $0.5$ \\
Feet slide & $\sum_i \|\mathbf{v}_{i,xy}\|_2^2\,\mathbb{I}(\text{contact}_i)$ & $-0.4$ \\
Feet at plane & $\sum_i (z_i-z_{\text{plane}})^2$ & $-0.1$ \\
Feet flat orientation & $\sum_i \|\mathbf{q}_{i,\text{err}}\|_2^2\,\mathbb{I}(\text{contact}_i)$ & $-0.4$ \\
Feet close  & $\exp(-d_{xy}^2/\sigma^2)$ & $0.4$ \\
Foothold penalty & $\sum_{i\in\{\mathrm{L},\mathrm{R}\}} \mathbb{I}(c_i=1)\mathbb{I}(h_i<h_{\mathrm{thr}})$ & $-2.0$ \\
Volume penetration & $\sum \max(0, z_{\text{terrain}}-z_{\text{point}})$ & $-5.0$ \\
Undesired contacts & $\sum_i \mathbb{I}(\|\mathbf{F}_i\|_2>F_{\text{thr}})$ & $-1.0$ \\
Joint torques & $\sum_i \tau_i^2$ & $-1.5e^{-7}$ \\
Joint acceleration & $\sum_i \ddot{\theta}_i^2$ & $-1.25e^{-7}$ \\
Joint velocity & $\sum_i \dot{\theta}_i^2$ & $-1.0e^{-4}$ \\
Action rate & $\|\mathbf{a}_t-\mathbf{a}_{t-1}\|_2^2$ & $-0.005$ \\
Energy & $\sum_i P_i^2$ & $-5.0e^{-5}$ \\
CoT penalty & $\sum_i |\tau_i \dot q_i| \,/\, (m g \|\mathbf{v}\|)$ & $-0.05$ \\
Joint pos limits & $\sum_i \mathrm{ReLU}(\theta_i-\theta_{\max})+\mathrm{ReLU}(\theta_{\min}-\theta_i)$ & $-1.0$ \\
Joint vel limits & $\sum_i \mathrm{ReLU}(|\dot{\theta}_i|-\dot{\theta}_{\max})$ & $-1.0$ \\
Torque limits & $\sum_i \mathrm{ReLU}(|\tau_i|/\tau_{\max}-0.8)$ & $-0.01$ \\

\midrule

AMP reward & $-\log\!\bigl(1 - D_\phi(\mathbf{s}_t)\bigr)$ & $0.25$ \\

\bottomrule
\end{tabularx}
\vspace{-0.3cm}
\end{table}

\subsection{Domain Randomization}
To improve the real-world generalization of the learned policy, we introduce moderate domain randomization during training across dynamics, depth perception, and command inputs. This includes small perturbations to physical properties such as mass and friction, staged noise applied to ray-cast depth observations, and mild variations in velocity and heading commands. Periodic external forces are also applied to enhance disturbance robustness. Together, these randomizations strengthen the policy’s stability in real-world settings, with the detailed configurations summarized in Table \ref{table:dr}.

\subsection{Curriculum Learning}
To improve the policy’s adaptability to complex terrains, we adopt a curriculum learning strategy. During training, the environment is divided into 10 levels, and the curriculum level is dynamically increased based on the achieved velocity reward. Higher levels correspond to larger commanded target velocities and progressively more challenging terrains. This design guides the policy to learn robust locomotion behaviors in a gradual and structured manner.

\begin{table}[!htbp]
\centering
\small
\caption{Parameters for Domain Randomization During Training}
\label{table:dr}
\begin{tabularx}{\columnwidth}{@{}>{\raggedright\arraybackslash}X >{\centering\arraybackslash}X@{}}
\toprule
\textbf{Term} & \textbf{Randomization Range} \\
\midrule
Static Friction & [0.1, 3.0] \\
Dynamic Friction & [0.1, 3.0] \\
Ground Restitution & [0.05, 0.5] \\
Base Mass Offset & [-5.0, 5.0] \\
$K_p$ Scale & [0.9, 1.1] \\
$K_d$ Scale & [0.9, 1.1] \\
Base Orientation Reset & [-0.1, 0.1] \\
Linear Velocity Reset & [-0.2, 0.2]\\
Angular Velocity Reset & [-0.2, 0.2] \\
Linear Velocity Perturbation & [-0.2, 0.2] \\
Initial Joint Positions & [-0.15, 0.15] \\
Initial Joint Velocities & [0.0, 0.1] \\
\midrule
Crop \& Resize & [18, 0, 16, 16] \\
Gaussian Noise & $\sigma = 0.02$, range [0.12, 1.2] \\
Disparity Artifact Synthesis & $P_{\mathrm{mask}} = 0.05\%$ \\
Depth Gaussian Blur& kernel size = $3 \times 3$, $\sigma = 1.0$\\
Depth Range Normalization & [0.0, 2.5] $\to$ [0.0, 1.0] \\
OOD Perturbation Probability & $P_{\mathrm{ood}} = 0.02\%$ \\
\bottomrule
\end{tabularx}
\vspace{-0.5cm}
\end{table}

\section{Experiments}
In this section, we evaluate the robustness and performance of our two-stage adaptive locomotion framework through extensive experiments in both simulation and real-world environments. The policy is tested across flat and complex terrains, including stairs, platforms, slopes, and discrete gaps, summarized in Table \ref{table:terrain}, using both walking and running gaits. All quantitative results are reported as the mean and standard deviation (mean ± std) over multiple evaluation runs. Our evaluation aims to answer the following three questions:
\begin{itemize}
    \item \textbf{Q1}: In Stage 1, can the adaptive gait locomotion policy learn natural walking and running gaits on flat terrain, while enabling smooth transitions between them?
    \item \textbf{Q2}: In Stage 2, can the depth perception policy branch with contrastive learning enable stable locomotion and terrain-aware motion adjustment on complex terrains through action fusion with the Stage 1 policy?
    \item \textbf{Q3}: Can the trained policy be reliably deployed on physical hardware in a zero-shot manner, without additional fine-tuning?
\end{itemize}

\begin{table}[!htbp]
\vspace{-0.1cm}
\centering
\fontsize{8pt}{10pt}\selectfont 
\caption{Terrain Types and Corresponding Parameter Ranges}
\label{table:terrain}
\begin{tabularx}{\columnwidth}{@{}>{\raggedright\arraybackslash}p{2.2cm} >{\centering\arraybackslash}p{4.3cm} >{\centering\arraybackslash}p{1.3cm}@{}}
\toprule
\textbf{Terrain Type} & \textbf{Terrain Feature} & \textbf{Range} \\
\midrule
Flat Ground & Flat terrain with mild variations & 0.0--0.1m \\
Up Slope & Upward inclined terrain & 0.0--35.0$^\circ$ \\
Down Slope & Downward inclined terrain & 0.0--35.0$^\circ$ \\
Up Stairs & Standard stair ascent terrain & 0.0--0.2m \\
Down Stairs & Standard stair descent terrain & 0.0--0.2m \\
High Up Stairs & Steep and tall stair ascent terrain & 0.0--0.45m \\
High Down Stairs & Steep and tall stair descent terrain & 0.0--0.45m \\
High Obstacle & Tall single-step obstacle terrain & 0.1--0.4m \\
Random Obstacle & Randomly distributed obstacle terrain & 0.1--0.3m \\
\bottomrule
\end{tabularx}
\vspace{-0.3cm}
\end{table}

\subsection{Evaluation of the Adaptive Gait Locomotion Policy}
We compared the proposed Adaptive Gait Locomotion Policy with three baseline methods: \textbf{(1) Mixed AMP}: which merges walking and running datasets to train a single AMP network; \textbf{(2) Sel AMP}: which selects walking or running references based on the commanded velocity to train a single AMP network; \textbf{(3) w/o MoE}: a single actor-critic network without using the MoE architecture.

To evaluate the performance of the Stage-1 adaptive gait policy on flat terrain, the robot receives distance-based velocity commands while navigating toward target points. When far from the target, the commanded velocity increases in the range of 1 to 2.5 m/s to induce a running gait, while near the target the velocity gradually decreases in the range of 0 to 1 m/s to produce a walking gait. This setup examines whether the policy can generate natural walking and running gaits and achieve smooth gait transitions on flat ground.

Locomotion performance was evaluated using four metrics: (1) \textbf{Velocity Tracking Score (Vel. Score.)}: An averaged exponential kernel score computed from the squared error between the commanded and actual base linear velocities in the xy-plane. (2) \textbf{Success Rate (Succ.)}: The percentage of trials in which the robot successfully reaches the specified target point; (3) \textbf{Average Traversal Ratio (Dist.)}: The fraction of the target distance completed before reaching the goal or termination; (4) \textbf{Side-view Gait Sequencess}: Visual inspection of the robot’s side-view motion sequences to assess the naturalness and smoothness of the walking and running gaits.

As shown in Table \ref{table:Flat Ground}, CoRe-MoE achieves the highest velocity tracking score, success rate, and traversal ratio. Moreover, the relatively small standard deviations indicate that the proposed policy maintains consistent performance across different evaluations. 

\begin{figure}[t]
    \centering
    \includegraphics[width=1\linewidth]{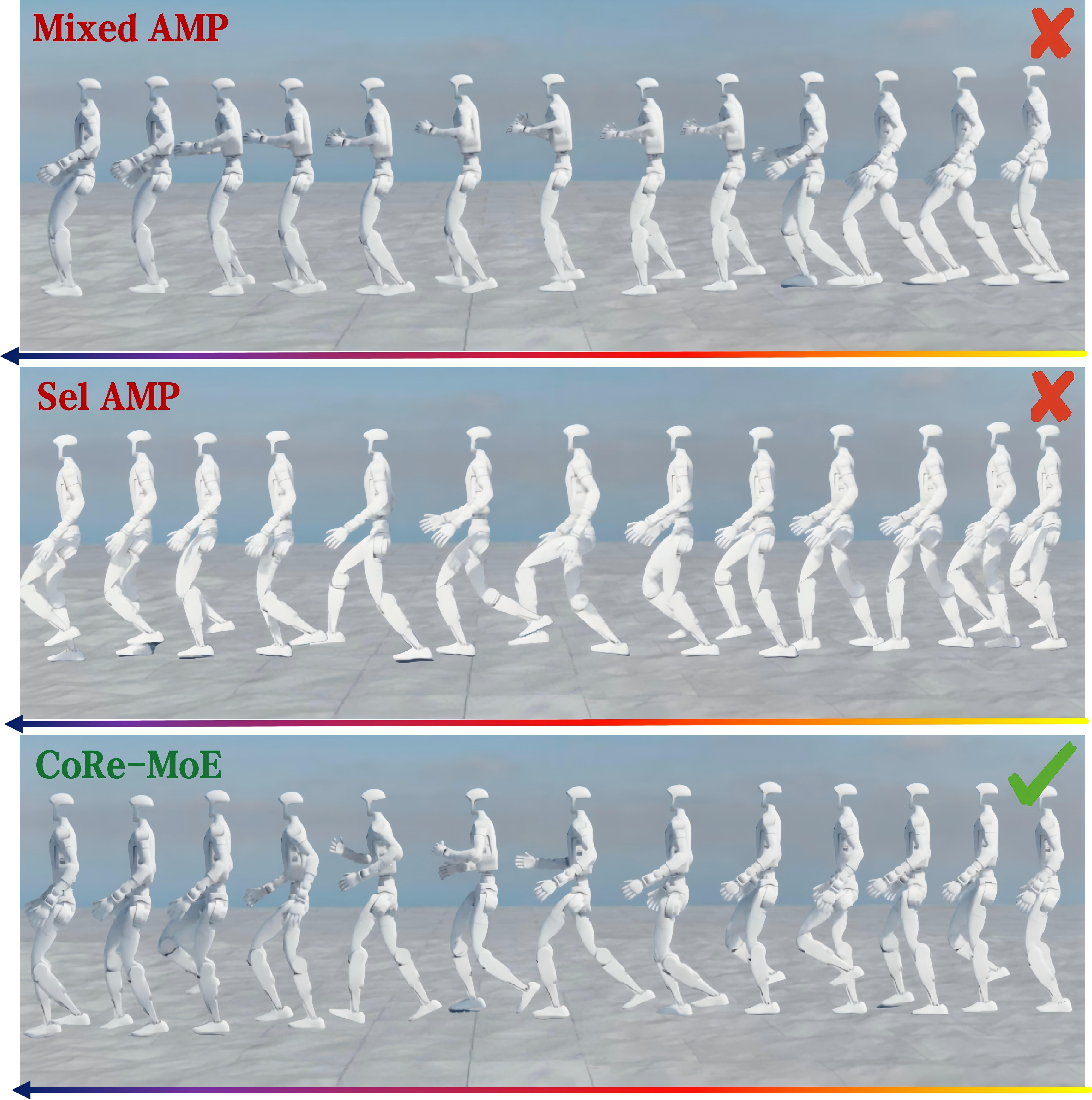}
    \caption{Side-view gait sequence visualization results. CoRe-MoE exhibits the most natural gait.}
    \vspace{-0.5cm}
    \label{figure:walk_run}
\end{figure}

The side-view gait sequences in Figure \ref{figure:walk_run} further reveal the differences among the policies. For Mixed AMP, merging walking and running data into a single network causes it to learn an intermediate shuffling gait between walking and running, resulting in unnatural motion. Sel-AMP, although selecting the corresponding reference data based on the commanded velocity, often incorrectly increases stride length under high-speed commands, producing an unnatural over-stride behavior. This leads to unstable mixed gait characteristics and prevents the emergence of clear walking or running patterns. In addition, both baseline methods exhibit unnatural arm movements. 

In contrast, CoRe-MoE employs separate AMP experts for walking and running and dynamically selects them based on the commanded velocity, enabling the policy to learn distinct walking and running gaits at low and high speeds. Moreover, modeling each gait with its own expert results in more natural and coherent arm swings, yielding significantly better gait continuity and adaptability compared to the baselines. We strongly recommend watching the demonstration videos on our project homepage for a more intuitive understanding.

\begin{table}[h]
    \centering
    \renewcommand{\arraystretch}{1.0}
    \setlength{\tabcolsep}{4pt}
    \fontsize{8.5pt}{10pt}\selectfont
    \caption{Performance Comparison of Adaptive Gait Policy and Baseline Methods on Flat Ground}
    \label{table:Flat Ground}
    \begin{tabularx}{\linewidth}{
        >{\centering\arraybackslash}X
        >{\centering\arraybackslash}X
        >{\centering\arraybackslash}X
        >{\centering\arraybackslash}X
    }
        \toprule
        \textbf{Method} & \textbf{Vel. Score.} & \textbf{Succ.} & \textbf{Dist.} \\
        \midrule
        Mixed AMP
        & $0.796 \pm 0.008$
        & $96.12 \pm 0.37\%$
        & $97.60 \pm 0.29\%$ \\

        Sel AMP
        & $0.812 \pm 0.015$
        & $54.20 \pm 1.84\%$
        & $54.50 \pm 1.67\%$ \\

        w/o MoE
        & $0.817 \pm 0.007$
        & $97.06 \pm 0.31\%$
        & $98.41 \pm 0.24\%$ \\

        \textbf{CoRe-MoE}
        & $\mathbf{0.853 \pm 0.004}$
        & $\mathbf{99.13 \pm 0.15\%}$
        & $\mathbf{99.85 \pm 0.08\%}$ \\
        \bottomrule
    \end{tabularx}
    \vspace{-0.3cm}
\end{table}

\subsection{Evaluation of the Adaptive Multi-Terrain Policy}
We compare the proposed Adaptive Multi-Terrain Policy with five baseline variants to analyze the contribution of each key component:\textbf{(1) w/o MoE}: removing the MoE architecture; \textbf{(2) Masked Expert}: zeroing out the outputs of activated experts; \textbf{(3) w/o Fusion}: removing the second-stage policy branch and directly fine-tuning the Stage 1 policy on multi-terrain tasks; \textbf{(4) w/o SwAV}: removing SwAV-based contrastive learning; \textbf{(5) SimCLR}: using instance-level SimCLR contrastive learning instead of SwAV.

\begin{figure}[b]
    \vspace{-0.5cm}
    \centering
    \includegraphics[width=1\linewidth]{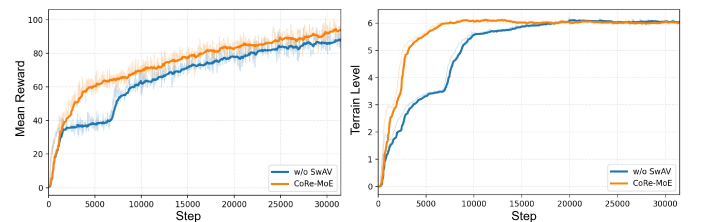}
    \caption{Evolution of Mean Reward and Terrain Curriculum Level Across Training Iterations}
    \label{figure:Iteration}
\end{figure}

To evaluate the policy under complex terrain conditions, the robot is deployed on diverse terrains including stairs, slopes and obstacle. We retain the same distance-dependent velocity setting used in Stage 1 to create mixed walking–running scenarios. The key distinction is that this stage focuses on assessing stability under terrain-induced disturbances and determining whether the policy can leverage the second-stage policy branch and the contrastively trained MoE gating mechanism to generate terrain-aware corrective actions for robust traversal over challenging terrains.

Locomotion performance was quantified using multiple metrics: \textbf{(1) Success rate (Succ.)}: The percentage of trials in which the robot successfully reached the target point; \textbf{(2) Average Traversal Ratio (Dist.)}: The fraction of the target distance completed before reaching the goal or termination; \textbf{(3) Convergence time (Conv. Time)}: The number of training iterations required for the policy to reach stable performance; \textbf{(4) Discriminability of terrain features}: evaluated via t-SNE \cite{t-SNE} visualization of MoE gating outputs to assess how well the embeddings differentiate between different terrain types.

\begin{table*}[t]
  \centering  
  \caption{Ablation Study of CoRe-MoE with Multi-Terrain Locomotion Performance Analysis. All results are reported as mean values over evaluation episodes. Standard deviations are omitted for clarity.}
  \renewcommand{\arraystretch}{1.3}
  \setlength{\tabcolsep}{3.0pt}
  \fontsize{6.8pt}{6.0pt}\selectfont
  \label{table:Multi-Terrain}
  \begin{tabular}{c*{9}{cc}}
    \toprule
    \multirow{2}{*}{\textbf{Method}} 
      & \multicolumn{2}{c}{\textbf{Flat Ground}}
      & \multicolumn{2}{c}{\textbf{Up Slope}}
      & \multicolumn{2}{c}{\textbf{Down Slope}}
      & \multicolumn{2}{c}{\textbf{Up Stairs}}
      & \multicolumn{2}{c}{\textbf{Down Stairs}}
      & \multicolumn{2}{c}{\textbf{High Up Stairs}}
      & \multicolumn{2}{c}{\textbf{High Down Stairs}}
      & \multicolumn{2}{c}{\textbf{High Obstacle}}
      & \multicolumn{2}{c}{\textbf{Random Obstacle}} \\
    \cmidrule(lr){2-3}
    \cmidrule(lr){4-5}
    \cmidrule(lr){6-7}
    \cmidrule(lr){8-9}
    \cmidrule(lr){10-11}
    \cmidrule(lr){12-13}
    \cmidrule(lr){14-15}
    \cmidrule(lr){16-17}
    \cmidrule(lr){18-19}
      & \multicolumn{2}{c}{Succ. \hspace{2.0pt} Dist.}
      & \multicolumn{2}{c}{Succ. \hspace{2.0pt} Dist.}
      & \multicolumn{2}{c}{Succ. \hspace{2.0pt} Dist.}
      & \multicolumn{2}{c}{Succ. \hspace{2.0pt} Dist.}
      & \multicolumn{2}{c}{Succ. \hspace{2.0pt} Dist.}
      & \multicolumn{2}{c}{Succ. \hspace{2.0pt} Dist.}
      & \multicolumn{2}{c}{Succ. \hspace{2.0pt} Dist.}
      & \multicolumn{2}{c}{Succ. \hspace{2.0pt} Dist.}
      & \multicolumn{2}{c}{Succ. \hspace{2.0pt} Dist.} \\
    \midrule
  w/o MoE & 
     \multicolumn{2}{c}{97.6\% \hspace{2.0pt} 99.5\%} & \multicolumn{2}{c}{76.4\% \hspace{2.0pt} 82.6\%} & \multicolumn{2}{c}{79.1\% \hspace{2.0pt} 86.1\%} & \multicolumn{2}{c}{35.6\% \hspace{2.0pt} 54.7\%} & \multicolumn{2}{c}{40.7\% \hspace{2.0pt} 63.2\%} & \multicolumn{2}{c}{12.2\% \hspace{2.0pt} 30.1\%} & \multicolumn{2}{c}{16.7\% \hspace{2.0pt} 38.9\%} & 
     \multicolumn{2}{c}{23.5\% \hspace{2.0pt} 41.3\%} & 
     \multicolumn{2}{c}{29.8\% \hspace{2.0pt} 46.2\%} \\
  Masked Expert & 
     \multicolumn{2}{c}{19.2\% \hspace{2.0pt} 32.6\%} & \multicolumn{2}{c}{4.9\%  \hspace{2.0pt} 19.3\%} & \multicolumn{2}{c}{15.6\% \hspace{2.0pt} 27.4\%} & \multicolumn{2}{c}{2.1\%  \hspace{2.0pt} 6.7\%} & \multicolumn{2}{c}{9.4\%  \hspace{2.0pt} 24.7\%} & \multicolumn{2}{c}{0.2\%  \hspace{2.0pt} 2.3\%} & \multicolumn{2}{c}{4.3\%  \hspace{2.0pt} 15.6\%} & 
     \multicolumn{2}{c}{7.6\%  \hspace{2.0pt} 21.0\%} & 
     \multicolumn{2}{c}{3.5\%  \hspace{2.0pt} 12.5\%} \\
  w/o Fusion & 
     \multicolumn{2}{c}{98.7\% \hspace{2.0pt} 99.8\%} & \multicolumn{2}{c}{81.2\% \hspace{2.0pt} 87.6\%} & \multicolumn{2}{c}{86.3\% \hspace{2.0pt} 93.7\%} & \multicolumn{2}{c}{36.3\% \hspace{2.0pt} 58.6\%} & \multicolumn{2}{c}{51.8\% \hspace{2.0pt} 70.4\%} & \multicolumn{2}{c}{9.4\%  \hspace{2.0pt} 24.8\%} & \multicolumn{2}{c}{17.5\% \hspace{2.0pt} 42.3\%} & 
     \multicolumn{2}{c}{26.1\% \hspace{2.0pt} 43.4\%} & 
     \multicolumn{2}{c}{35.4\% \hspace{2.0pt} 58.5\%} \\
  w/o SwAV & 
    \multicolumn{2}{c}{99.5\% \hspace{2.0pt} 99.9\%} & \multicolumn{2}{c}{96.4\% \hspace{2.0pt} 98.5\%} & \multicolumn{2}{c}{98.2\% \hspace{2.0pt} 99.4\%} & \multicolumn{2}{c}{84.6\% \hspace{2.0pt} 92.3\%} & \multicolumn{2}{c}{86.3\% \hspace{2.0pt} 93.6\%} & \multicolumn{2}{c}{72.3\% \hspace{2.0pt} 85.1\%} & \multicolumn{2}{c}{79.5\% \hspace{2.0pt} 87.2\%} & 
    \multicolumn{2}{c}{90.2\% \hspace{2.0pt} 96.9\%} & 
    \multicolumn{2}{c}{81.5\% \hspace{2.0pt} 89.6\%} \\
  SimCLR & 
    \multicolumn{2}{c}{99.7\% \hspace{2.0pt} 99.9\%} & \multicolumn{2}{c}{97.3\% \hspace{2.0pt} 99.1\%} & \multicolumn{2}{c}{98.6\% \hspace{2.0pt} 99.7\%} & \multicolumn{2}{c}{87.6\% \hspace{2.0pt} 94.3\%} & \multicolumn{2}{c}{89.4\% \hspace{2.0pt} 94.2\%} & \multicolumn{2}{c}{76.8\% \hspace{2.0pt} 89.3\%} & \multicolumn{2}{c}{81.6\% \hspace{2.0pt} 90.3\%} & 
    \multicolumn{2}{c}{93.4\% \hspace{2.0pt} 97.5\%} & 
    \multicolumn{2}{c}{83.6\% \hspace{2.0pt} 91.4\%} \\
  \textbf{CoRe-MoE} & 
     \multicolumn{2}{c}{\textbf{99.9\%} \hspace{2.0pt} \textbf{99.9\%}} & \multicolumn{2}{c}{\textbf{98.7\%} \hspace{2.0pt} \textbf{99.4\%}} & \multicolumn{2}{c}{\textbf{98.9\%} \hspace{2.0pt} \textbf{99.9\%}} & \multicolumn{2}{c}{\textbf{94.3\%} \hspace{2.0pt} \textbf{96.8\%}} & \multicolumn{2}{c}{\textbf{96.5\%} \hspace{2.0pt} \textbf{97.9\%}} & \multicolumn{2}{c}{\textbf{86.4\%} \hspace{2.0pt} \textbf{91.6\%}} & \multicolumn{2}{c}{\textbf{87.1\%} \hspace{2.0pt} \textbf{93.7\%}} & 
     \multicolumn{2}{c}{\textbf{97.6\%} \hspace{2.0pt} \textbf{98.7\%}} & 
     \multicolumn{2}{c}{\textbf{89.3\%} \hspace{2.0pt} \textbf{92.8\%}} \\
    \bottomrule
  \end{tabular}
  \vspace{-0.5cm}
\end{table*}

\begin{figure}[t]
    \centering
    \includegraphics[width=1\linewidth]{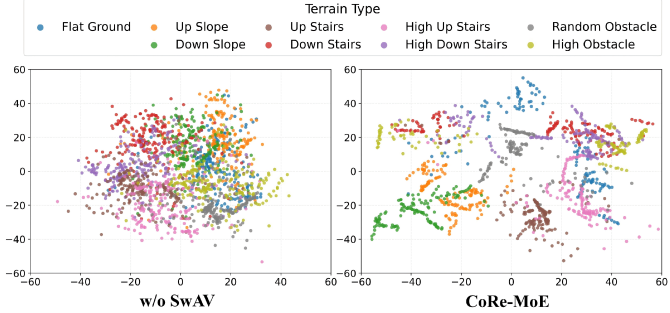}
    \caption{T-SNE visualization of MoE gating features across different terrains. The left figure (w/o SwAV) shows mixed and overlapping feature distributions, while the right figure (CoRe-MoE) presents more compact and clearly separated clusters.}
    \vspace{-0.5cm}
    \label{figure:t-SNE}
\end{figure}

As shown in Table \ref{table:Multi-Terrain} and Figure \ref{figure:Iteration}, (1) w/o MoE exhibits a significant drop in success rate and traversal ratio across most terrains, indicating that MoE plays a central role in expert decomposition and policy specialization; (2) Masked Expert severely degrades performance by zeroing out activated expert outputs, further validating the importance of correct expert selection for terrain-aware control; (3) w/o Fusion shows reduced stability in complex multi-terrain scenarios and performs reasonably only on structurally simple terrains such as stairs, demonstrating that this fusion mechanism is essential for cross-terrain generalization; (4) w/o SwAV leads to lower success rate and slower convergence, suggesting that SwAV contrastive learning helps the MoE gating network better distinguish terrain features and accelerate policy optimization; (5) SimCLR performs better than w/o SwAV on most terrains but still underperforms CoRe-MoE overall, indicating that although instance-level contrastive learning improves representation quality, it lacks structured terrain semantic modeling capability and cannot form effective cluster-level alignment between terrains and experts, thereby limiting its effectiveness in MoE expert selection.

In contrast, CoRe-MoE consistently achieves the highest success rates and traversal ratios across all evaluated terrains, demonstrating superior robustness and generalization capability. Nevertheless, a small number of failures still occur in the most challenging scenarios, particularly on High Up Stairs and Random Obstacle terrains. We attribute these failures primarily to the abrupt terrain height variations and limited foothold margins in highly irregular environments, where small perception or foot-placement errors can accumulate and eventually lead to loss of balance.

Figure \ref{figure:t-SNE} presents t-SNE visualizations of MoE gating outputs across different terrain types. We specifically conducted this experiment to verify the effect of SwAV contrastive learning on terrain feature encoding. The embeddings with SwAV show clearly separated clusters for each terrain type, while embeddings without SwAV exhibit more overlap, indicating reduced discriminability. These results demonstrate that contrastive learning enables the gating network to more effectively distinguish terrain features and activate the corresponding experts for terrain-aware motion adjustments.

Overall, the proposed CoRe-MoE demonstrates superior robustness and stability in multi-terrain locomotion tasks, and achieves more reliable locomotion performance under complex terrain conditions.

\subsection{Comparison with Prior Locomotion Methods}
To further evaluate the proposed method, we compared CoRe-MoE with representative humanoid locomotion frameworks, including Hiking in the Wild \cite{zhu2026hiking} and MoRE \cite{MORE}. For concise presentation, the reported stair, slope, and obstacle success rates are averaged over the corresponding terrain variants, including both ascending and descending cases as well as different difficulty levels.

As shown in Table \ref{tab:policy_comparison}, Hiking in the Wild trains separate walking and running policies and requires manual switching during deployment, while MoRE employs a single policy but still relies on manually specified gait motion modes. Such approaches struggle to continuously adapt locomotion behaviors according to motion demands, limiting both adaptability and energy efficiency across a wide range of velocities. For example, using only a walking policy becomes inefficient at high speeds, whereas relying solely on running introduces unnecessary energy expenditure at low speeds. Moreover, manually specified motion modes cannot always guarantee optimal gait selection under varying terrain and velocity conditions.

In contrast, CoRe-MoE achieves unified walking and running control within a single policy and automatically performs gait transitions according to commanded velocities. In addition to the simplified policy structure, CoRe-MoE achieves the best overall locomotion performance, obtaining the highest velocity tracking score and success rates across multiple terrain categories. Furthermore, CoRe-MoE achieves the lowest Cost of Transport (CoT), indicating superior energy efficiency during locomotion. This advantage stems from the automatic walk run transition mechanism, which enables the policy to employ energy-efficient walking at low speeds, switch to more suitable running behaviors at high speeds, and smoothly transition between the two regimes in intermediate velocity ranges. As a result, CoRe-MoE avoids the efficiency loss associated with relying on a single locomotion mode across all velocities. These results demonstrate that the proposed framework effectively combines unified gait generation, automatic gait transition, energy-efficient locomotion, and robust multi-terrain traversal within a single policy architecture.

\begin{table*}[t]
\centering
\caption{Comparison of Policy Structures and Multi-Terrain Locomotion Performance Across Methods.}
\label{tab:policy_comparison}
\renewcommand{\arraystretch}{1.1}
\setlength{\tabcolsep}{4pt}
\fontsize{7.5pt}{8.5pt}\selectfont
\begin{tabular*}{\textwidth}{@{\extracolsep{\fill}}cccccccccc}
\toprule
\textbf{Method} &
\textbf{Unified Policy} &
\textbf{Transition Mode} &
\textbf{Vel. Score.} &
\textbf{Flat Succ.} &
\textbf{Slope Succ.} &
\textbf{Stairs Succ.} &
\textbf{Obstacle Succ.} &
\textbf{CoT}\\
\midrule

Hiking in the Wild {\fontsize{7pt}{8pt}\selectfont(Walk)}
& \ding{55}
& Manual
& $0.85 \pm 0.02$
& $99.7 \pm 0.2\%$
& $76.8 \pm 0.8\%$
& $82.6 \pm 0.7\%$
& $84.9 \pm 0.6\%$
& $0.75 \pm 0.02$ \\

Hiking in the Wild {\fontsize{7pt}{8pt}\selectfont(Run)}
& \ding{55}
& Manual
& $0.62 \pm 0.03$
& $99.4 \pm 0.3\%$
& $67.9 \pm 1.1\%$
& $54.7 \pm 1.4\%$
& $78.5 \pm 0.9\%$ 
& $0.67 \pm 0.02$ \\

MoRE
& \ding{51}
& Manual
& $0.79 \pm 0.02$
& $97.6 \pm 0.4\%$
& $93.7 \pm 0.5\%$
& $90.3 \pm 0.6\%$
& $91.2 \pm 0.5\%$
& $0.82 \pm 0.03$ \\

\textbf{CoRe-MoE}
& \textbf{\ding{51}}
& \textbf{Automatic}
& $\mathbf{0.89 \pm 0.01}$
& $\mathbf{99.9 \pm 0.1\%}$
& $\mathbf{98.8 \pm 0.3\%}$
& $\mathbf{91.1 \pm 0.4\%}$
& $\mathbf{93.5 \pm 0.3\%}$
& $\mathbf{0.61 \pm 0.01}$ \\
\bottomrule
\end{tabular*}
\vspace{-0.1cm}
\end{table*}

\begin{figure*}
    \centering
    \includegraphics[width=1.0\linewidth]{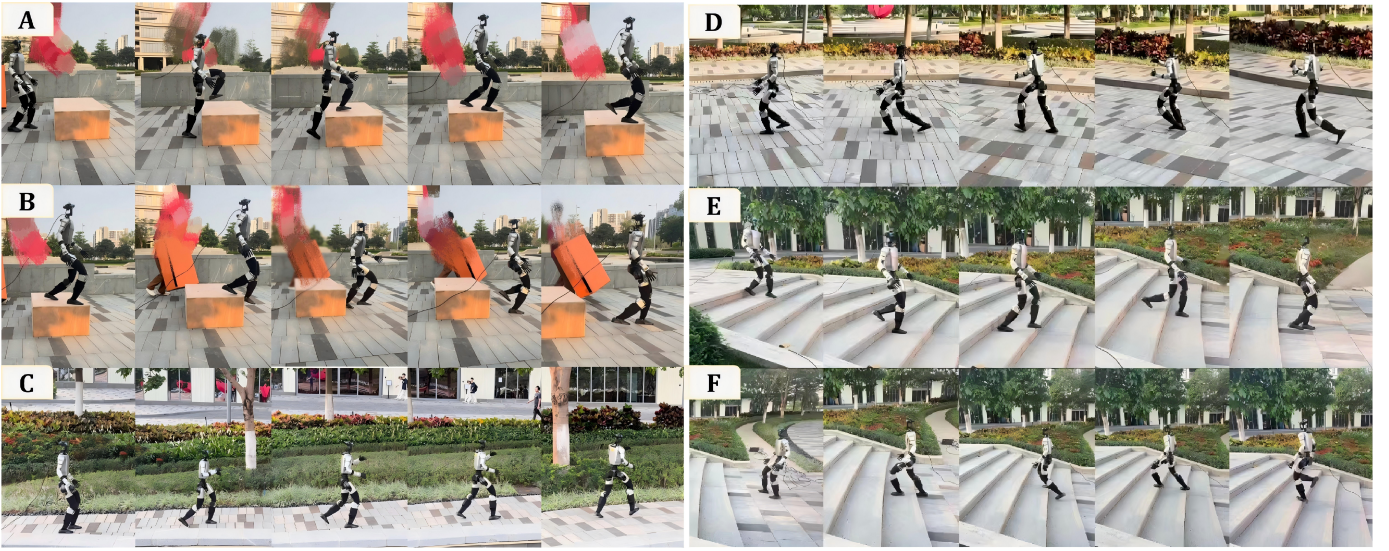}
\caption{The multi-terrain sequence demonstrates the robot’s continuous locomotion across six representative terrains, where A-F correspond to obstacle ascent, obstacle descent, slope ascent, flat ground, stair descent, and stair ascent. The robot smoothly transitions between different terrains while maintaining stable gait and reliable foothold control, demonstrating strong multi-terrain adaptability.}
\vspace{-0.4cm}
\label{figure:mutli-terrain}
\end{figure*}

\begin{figure}
    \centering
    \includegraphics[width=1\linewidth]{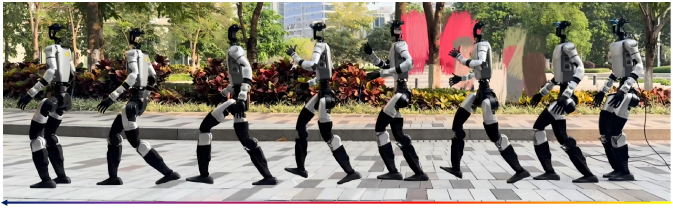}
    \caption{Side-view gait sequence of the humanoid robot in real world deployment, showing a continuous transition from walking to running and back to walking without any mode switching.}
    \vspace{-0.5cm}
    \label{figure:real_walk_run}
\end{figure}

\subsection{Real-world Experiment}
To validate the generalization capability of the proposed method CoRe-MoE on real robotic hardware, we directly deployed the trained policy onto the Unitree G1 humanoid robot without any additional fine-tuning. The system uses the onboard Intel RealSense D435i depth camera to capture real-time depth images, which are processed through lightweight geometric operations and fed into the policy network, enabling zero-shot transfer from simulation to the real robot.

All experiments are conducted in real outdoor environments, including uneven ground, natural steps, outdoor obstacles, natural slopes, and terrains with height discontinuities. Despite significant discrepancies in geometry, surface materials, and lighting conditions compared to simulation, the robot consistently maintains stable locomotion with accurate foothold placement.

As shown in Figure \ref{figure:real_walk_run}, the robot executes a continuous sequence consisting of walking, running, and returning to walking on flat ground. This experiment highlights the natural and coherent behavior of CoRe-MoE under varying velocity commands. Since the framework is trained as a single unified policy, the robot automatically adjusts its gait as speed increases or decreases, without relying on any manually triggered mode-switching mechanism. Even during high-speed motions, where the maximum outdoor forward velocity reaches 2.5 m/s, the robot exhibits stable and coordinated dynamics with no noticeable discontinuities.

\begin{figure}
    \centering
    \includegraphics[width=1\linewidth]{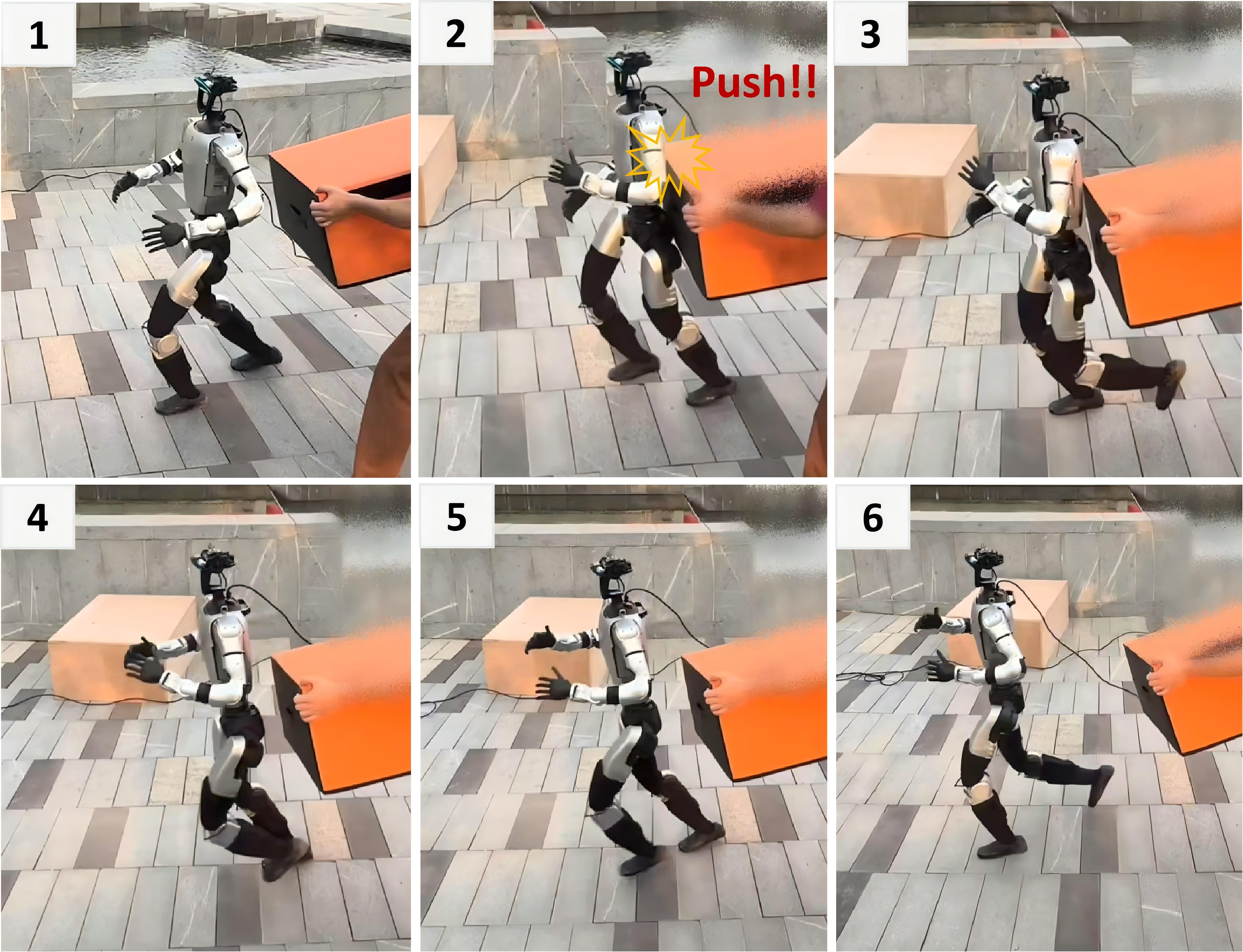}
    \caption{The robot is subjected to a sudden external impact while running, maintaining balance and quickly recovering stable locomotion.}
    \vspace{-0.7cm}
    \label{figure:robust}
\end{figure}

To further illustrate the advantage of unified gait control, we provide a comparison with the Hiking in the Wild baseline, as shown in Figure \ref{figure:cover}. The baseline trains walking and running policies separately, and switching between them must be performed manually during execution. Such manual switching often introduces brief discontinuities or posture inconsistencies at the transition points, resulting in a visibly fragmented motion sequence. In contrast, CoRe-MoE preserves natural, smooth, and dynamically consistent gait transitions throughout the entire sequence, including walking, running, returning to walking. The unified policy adaptively regulates stride length, center-of-mass motion, and whole-body coordination, demonstrating superior continuity compared with methods that depend on switching between multiple independent controllers.

Furthermore, in multi-terrain traversal tasks, the robot exhibits strong adaptability across different terrain types. The temporal sequence in Figure \ref{figure:mutli-terrain} illustrates the robot traversing uneven ground, natural steps, and slopes in a continuous manner. Throughout the process, the policy dynamically adjusts step length, foothold placement, and body posture in response to terrain variations, while maintaining stability under height discontinuities and varying friction conditions. These results demonstrate that the proposed method enables robust and coherent locomotion across diverse and challenging terrain compositions.

To further evaluate the system's robustness in real-world conditions, we conducted an impact disturbance test. During level-ground locomotion, the robot was subjected to a sudden external impact while running, as shown in Figure \ref{figure:robust}. Despite the disturbance occurring under high-speed motion, the robot maintained balance and quickly recovered stable locomotion. This experiment demonstrates that the proposed method provides strong stability and recovery capability under external perturbations during dynamic gait execution.

\section{Conclusions}
In this work, we presented a humanoid locomotion framework CoRe-MoE that integrates visual perception, weighted action refinement, and contrastively enhanced mixture-of-experts. The proposed method enables terrain-aware corrections and improves expert specialization through SwAV, allowing smooth and adaptive transitions between walking and running under varying velocity commands. Extensive simulations show that our approach outperforms baseline policies in success rate, travel distance, and multi-terrain adaptability. The trained policy is deployed onto a Unitree G1 robot without any fine-tuning, achieving stable zero-shot sim-to-real transfer using only the onboard depth camera. In both indoor and outdoor experiments, the robot demonstrates strong robustness, natural gait transitions, and resilience to external disturbances. Future work will extend this framework to whole-body coordination for more dynamic and human-like locomotion behaviors.

\bibliographystyle{unsrt}
\bibliography{ref}

\end{document}